\documentclass[lettersize,journal]{IEEEtran}
\usepackage{amsmath,amsfonts}
\usepackage{array}
\usepackage{algorithmicx,algorithm}
\usepackage[caption=false,font=normalsize,labelfont=sf,textfont=sf]{subfig}
\usepackage[noend]{algpseudocode}
\usepackage{textcomp}
\usepackage{stfloats}
\usepackage{url}
\usepackage{verbatim}
\usepackage{graphicx}
\usepackage{cite}
\usepackage{hyperref}
\usepackage{booktabs}
\usepackage{blindtext}
\usepackage{tabularx}
\usepackage{multirow}
\usepackage{pifont}
\usepackage[capitalize]{cleveref}
\usepackage{bbold}
\usepackage{amssymb}
\usepackage{indentfirst}
\usepackage{CJK}
\usepackage{autobreak}
\usepackage{tikz}
\usepackage{epstopdf}
\usepackage{diagbox}
\usepackage{epstopdf}
\usepackage{wasysym}
\usepackage{balance}
\usepackage{makecell}
\usepackage[normalem]{ulem}
\crefname{section}{Sec.}{Secs.}
\Crefname{section}{Section}{Sections}
\Crefname{table}{Table}{Tables}
\crefname{table}{Tab.}{Tabs.}
\hypersetup{hypertex=true,
            colorlinks=true,
            linkcolor=blue,
            anchorcolor=blue,
            citecolor=blue}

\begin{document}


\title{A Class-wise Non-salient Region Generalized Framework for Video Semantic Segmentation}

\author{Yuhang Zhang, Shishun Tian, Muxin Liao, Zhengyu Zhang, Wenbin Zou, Chen Xu

\thanks{Y. Zhang, S. Tian, M. Liao, are with Guangdong Key Laboratory of Intelligent Information Processing, College of Electronics and Information Engineering, Shenzhen University, Shenzhen, 518060, China (e-mail: zhangyuhang2019@email.szu.edu.cn, stian@szu.edu.cn, liaomuxin2020@email.szu.edu.cn).}
\thanks{Z. Zhang is with the Univ Rennes, INSA Rennes, CNRS, IETR - UMR 6164, F-35000 Rennes, France (email: zhengyu.zhang@insa-rennes.fr).}
\thanks{W. Zou is with Guangdong Key Laboratory of Intelligent Information Processing, Shenzhen Key Laboratory of Advanced Machine Learning and Applications, Institute of Artificial Intelligence and Advanced Communication, College of Electronics and Information Engineering, Shenzhen University, Shenzhen, 518060, China (e-mail: wzou@szu.edu.cn). (Corresponding author: Wenbin Zou.).}
\thanks{C. Xu is with the College of Mathematics and Statistics, Shenzhen University, Shenzhen 518060, China (e-mail: xuchen@szu.edu.cn).}}
\markboth{Journal of \LaTeX\ Class Files,~Vol.~14, No.~8, August~2021}%
{Shell \MakeLowercase{\textit{et al.}}: A Sample Article Using IEEEtran.cls for IEEE Journals}


\maketitle

\begin{abstract}
Video semantic segmentation (VSS) is beneficial for dealing with dynamic scenes due to the continuous property of the real-world environment. On the one hand, some methods alleviate the predicted inconsistent problem between continuous frames. On the other hand, other methods employ the previous frame as the prior information to assist in segmenting the current frame. Although the previous methods achieve superior performances on the independent and identically distributed (i.i.d) data, they can not generalize well on other unseen domains. Thus, we explore a new task, the video generalizable semantic segmentation (VGSS) task that considers both continuous frames and domain generalization. In this paper, we propose a class-wise non-salient region generalized (CNSG) framework for the VGSS task. Concretely, we first define the class-wise non-salient feature, which describes features of the class-wise non-salient region that carry more generalizable information. Then, we propose a class-wise non-salient feature reasoning strategy to select and enhance the most generalized channels adaptively. Finally, we propose an inter-frame non-salient centroid alignment loss to alleviate the predicted inconsistent problem in the VGSS task. We also extend our video-based framework to the image-based generalizable semantic segmentation (IGSS) task. Experiments demonstrate that our CNSG framework yields significant improvement in the VGSS and IGSS tasks.

\end{abstract}

\begin{IEEEkeywords}
Semantic segmentation, Video domain generalization, Non-salient region, Class-wise relationship reasoning, Domain-invariant feature
\end{IEEEkeywords}
\section{Introduction}
\label{Intro}
Semantic segmentation, applied in many applications such as robotics and medicinal diagnosis, has made significant progress due to the development of deep learning technology, which aims to give an object class for each pixel of an image. However, image-based semantic segmentation can not utilize the results of previous frames as the prior information to assist the segmentation of the current frame. Thus, some researchers studied the video-based semantic segmentation (VSS) task, which fully exploited the information of continuous temporal frames. These studies can be grouped into inter-frame fusion and inter-frame consistency to obtain a more accurate model. Although these approaches considered the temporal relationship, they suffer from performance degradation in environments with diverse domain styles, called the distribution shift problem. To tackle this problem, the number of scenarios that need to be generalized well is the key to selecting the appropriate technology. 

\par Unsupervised domain adaptation (UDA) is a preferred technology to handle the distribution shift problem for single scenes where the goal is to transfer the knowledge from the source domain to the target domain and achieve remarkable performance on the target domain. The source domain with annotations and the target domain without annotations are simultaneously used in the training stage. The UDA technology for semantic segmentation can be roughly divided into domain distribution alignment and self-training process. Domain distribution alignment focuses on reducing the domain gap by alignment strategies at the image, feature, and output levels. Self-training process selects confident pseudo-labels as the supervised labels of the target domain. These UDA methods have two challenges. First, only one scene can be adapted by the UDA methods. Second, the target data should be obtained for training which is not always feasible in the actual applications.
\begin{figure}[t]
\centering
\includegraphics[width=0.48\textwidth]{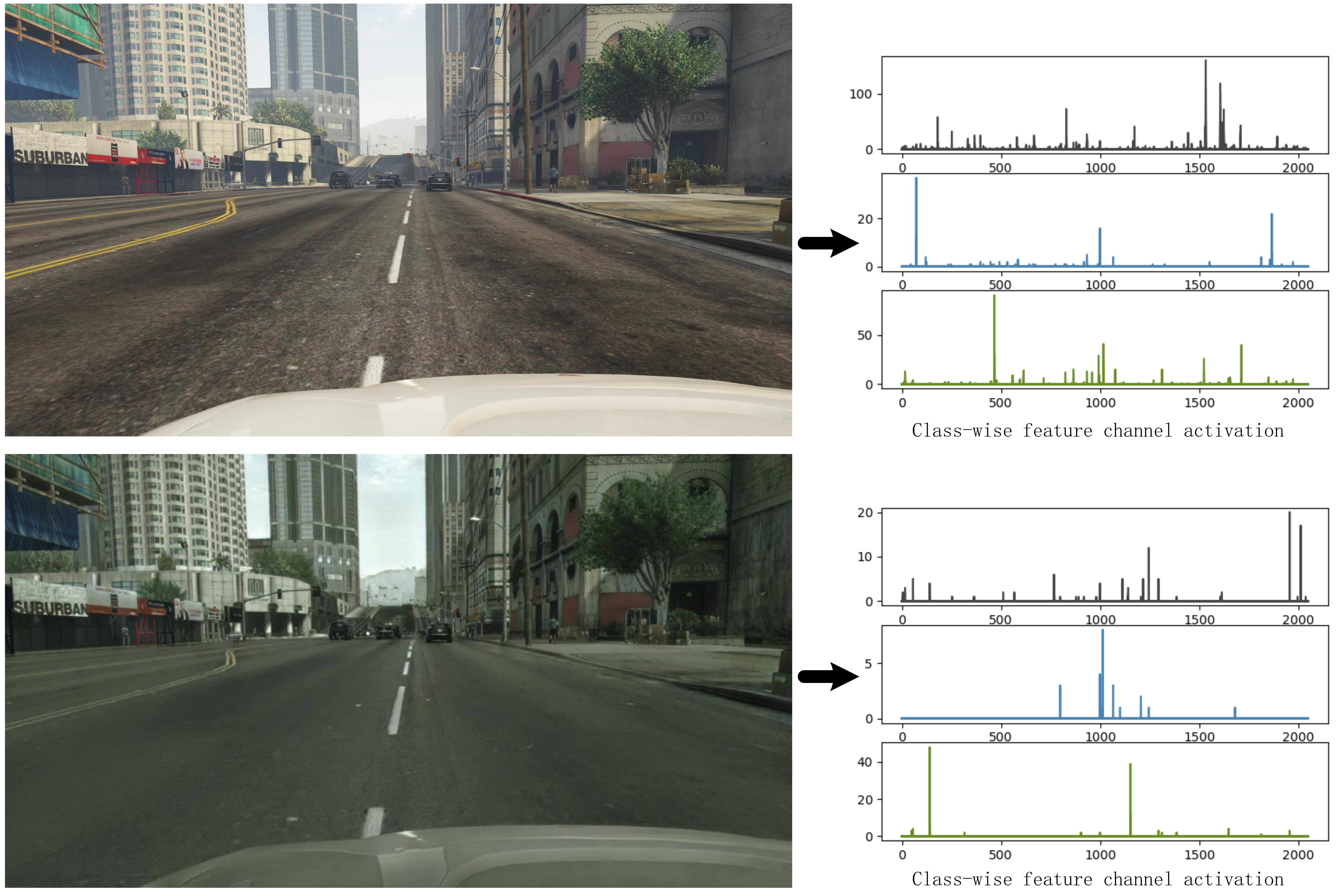}
\caption{The class-wise feature channel activation in images with different styles (GTAV \cite{richter2016playing} style, Cityscapes \cite{cordts2016cityscapes} style), where the shown classes are the road, sky, and vegetation from top to bottom.}
\label{Fig:FCD}
\end{figure}

\par Domain generalization (DG) is another technology to tackle the distribution shift problem for complex and multiple scenes, which is more practical than the UDA technology because DG can adapt to more scenes with diverse domains and can not need the target data in the training period. The purpose of the DG methods is to generalize well on unseen domains. Despite DG for image-based semantic segmentation (IGSS) got several explorations, there is no research touching the DG for video-based semantic segmentation. The video generalizable semantic segmentation (VGSS) task considers both temporal frames and domain generalizability. Compared with the IGSS task, the VGSS task is in line with the continuous attribute of the real world, which is significant for more accurate and robust prediction. Thus, we are devoted to designing a learning framework for the VGSS task. Three critical observations inspired us to devise a novel learning framework for the VGSS task, which are shown as follows.

\begin{figure}[t]
\centering
\includegraphics[width=0.48\textwidth]{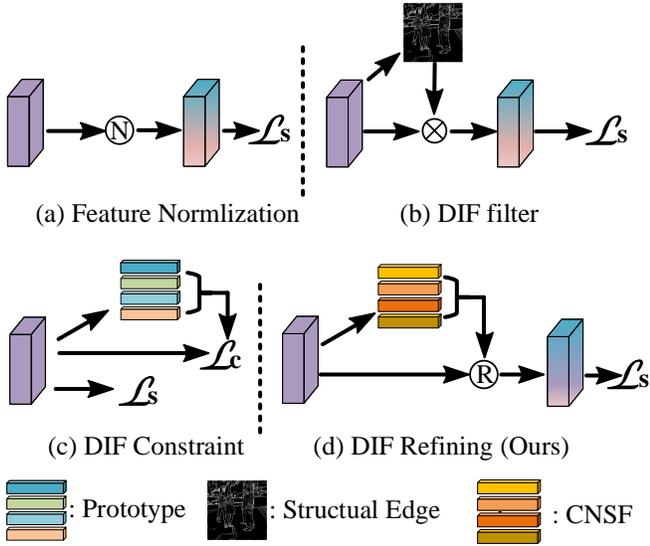}
\caption{Different ways to utilize domain-invariant features (DIF). \textcircled{\scriptsize N}, \textcircled{$\times$}, and \textcircled{\scriptsize R} refer to normalization, element-wise multiplication, and refining, respectively. CNSF, a type of DIF, represents the proposed class-wise non-salient feature.}
\label{Fig:GFR}
\end{figure}

\begin{enumerate}
\item \textbf{Feature channel activations of the same classes in different domains have a large gap.} Fig \ref{Fig:FCD} shows the class-wise feature channel activations of the last layer in two images with the same contents but different styles. Three classes are shown for simplicity. A large gap between diverse domains in channels means that the model can perceive the domain information from the training images rather than only the semantic information. The style information affects the class channel distribution, which may result in classified confusion. 

\item \textbf{Domain-invariant features (DIF) improve the generalizability of the model}, which are stable and do not change with styles. How to get and utilize DIF are two main challenges in handling the domain shift problem. As shown in Fig \ref{Fig:GFR}, the existing works mainly focus on DIF selection and DIF constraint. One type of DIF selection shown in Fig \ref{Fig:GFR} (a) aims to process all deep features and normalize them, such as feature normalization and whitening \cite{pan2018two, choi2021robustnet}. Another type of DIF selection shown in Fig \ref{Fig:GFR} (b) forces neural networks to pay more attention to DIF, such as structural edges \cite{shi2020informative} and features with small gradients \cite{huang2020self}. Based on DIF constraints, some methods extract class prototypes (the centroid of the class feature) and perform distribution alignment between the class feature and the class prototype \cite{liu2021bapa, lu2022bidirectional}, as shown in Fig \ref{Fig:GFR} (c). Different from these two types of approaches, we explore a novel form (see Fig \ref{Fig:GFR} (d)), which treats DIF as the prior information to attend the training and assists the model in adaptively refining features according to the current data.
\item \textbf{Predicted inconsistency between the adjacent temporal frames deteriorates generalization and accuracy.} Almost all video tasks of computer vision suffer from the predicted inconsistent problem \cite{li2021video}, which may lead to a performance decrease. The previous  VSS studies alleviate this issue under the condition of i.i.d data but may not be kindly applied to the VGSS task. Inter-frame domain-invariant feature constraint may be a worthful way to alleviate the predicted inconsistent problem in the distribution shift setting.
\end{enumerate}

\par Based on the three key observations mentioned in the previous paragraph, we propose a class-wise non-salient region generalized (CNSG) framework for the VGSS task. Specifically, we first define the class-wise non-salient feature, which describes features of the class-wise non-salient region that carry more generalizable information. Then, we propose a class-wise generalized feature reasoning module to select and enhance the most generalized channels adaptively. Finally, we propose an inter-frame non-salient centroid alignment loss to reduce the gap between the class-wise non-salient centroids of two adjacent frames. In summary, the main contributions of this work could be concluded as follow. 

\begin{itemize}
\item We introduce a new task for semantic segmentation, namely video generalizable semantic segmentation (VGSS), which has not been studied in existing research to the best of our knowledge. Additionally, we propose a class-wise non-salient region generalized (CNSG) framework for the VGSS task, which considers both continuous frames and domain generalization.
\item We adopt a novel form to utilize the domain-invariant feature, which treats that as the prior information to assist the model in adaptively refining features according to the current data. Concretely, we propose a class-wise non-salient feature reasoning strategy, which explores this form from the perspective of the channel distribution gap between different domains.
\item We propose an inter-frame non-salient centroid alignment loss to cope with the predicted inconsistent problem between adjacent frames, which extract and align the class-wise non-salient centroid for better generalizability.
\item Extensive experiments on different datasets and backbones show that our CNSG framework achieves superior performance. Furthermore, we extend our proposed CNSG framework to the IGSS task, which also gets competitive results compared to the state-of-the-art methods.
\end{itemize}

\par The structures of this article can be organized as follows. Section \ref{Sec:Rel} provides the related research direction, including semantic segmentation, unsupervised domain adaptation, domain generalization, and class activation map. Section \ref{Sec:Met} introduces the CNSG framework in detail. Section \ref{Sec:Exp} presents extensive experiments and analyses the experimental results. Section \ref{Sec:Coc} concludes the main contributions of this work.

\section{Literature review}
\label{Sec:Rel}
\subsection{Semantic segmentation}
\par Semantic segmentation, as a type of significant computer vision task, predicts the semantic classes of each pixel in images. Semantic segmentation methods can be roughly parted into architecture design and richer context aggregation. For architecture design, the fully convolutional network (FCN) \cite{long2015fully} and U-Net \cite{ronneberger2015u} were proposed as the baseline of many existing sophisticated methods, which kept both coarse and refined information depending on the skip connection operation. HRNet \cite{sun2019deep} maintained a high-resolution feature map with semantically strong. Recently, SETR \cite{zheng2021rethinking} and SegFormer \cite{strudel2021segmenter} developed transformer-based architectures which converted the original segmentation form into a sequence-to-sequence prediction task. RepMLPNet \cite{ding2022repmlpnet} proposed a multi-layer-perceptron (MLP) block with three fully connected layers to capture local priors by locality injection. Some methods got richer context by multi-scale information fusion like DeepLab-series methods \cite{chen2017deeplab, chen2017rethinking} and PSPNet \cite{zhao2017pyramid}. Attention is also a hotspot for capturing long-range dependencies such as CCNet \cite{huang2019ccnet}, Non-local \cite{wang2018non}, Segnext \cite{guosegnext}. Due to the continuous attribute of the real world, video-based methods attract attention. The video-based methods can be mainly split into reducing the cost of per-frame computation and improving segmentation performance. To reduce the cost of per-frame computation, DFF \cite{zhu2017deep} calculated the optical flow between the keyframe and the current frame and got predicted results by wrapping operation. DVS \cite{xu2018dynamic} proposed a dynamic selection strategy to segment by keyframe or segmentation network dynamically. LLVSS \cite{li2018low} got predicted results by the fusion between the low-level features of the current frame and the high-level features of the keyframe. For improving segmentation performance, the keyframe usually serves as the prior information to refine the feature of the current frame. EFC \cite{ding2020every} joint learned the video segmentation and optical flow tasks to improve both mutually. STT \cite{li2021video} utilized Transformer-based architecture to balance accuracy and efficiency. TMA \cite{wang2021temporal} developed inter-frame self-attention to perceive the inter-frame relationship better. The VGSS task is more challenging than the video semantic segmentation task due to the various unseen domains. Consequently, improving segmentation performance is the primary goal of this paper.

\subsection{Unsupervised domain adaptation}
Unsupervised domain adaptation (UDA) aims to perform well on the target domain given the label source and unlabeled target domains. Domain distribution alignment and self-training strategy are two types of general methods for UDA semantic segmentation. Domain distribution alignment parts image-level \cite{yang2020fda,tranheden2021dacs}, feature-level \cite{zhang2019category,wang2020differential}, and output-level \cite{liao2022exploring, zou2022dual, zhang2023hybrid} distribution alignment. Self-training strategy supervises unlabeled target data by the pseudo-label. BDL \cite{li2019bidirectional} used the max probability threshold to filter target pixels with confident prediction. ProDA \cite{zhang2021prototypical} utilized class prototypes to rectify the pseudo-label. Moreover, the prototype constraint shown in Fig \ref{Fig:GFR} (b) is effective and is used in many UDA methods, where the class prototype is the class feature centroid and can be seen as a type of domain-invariant feature. ProCA \cite{jiang2022prototypical} presented prototypical contrast adaptation, which pulled closer to the pixel and its corresponding class prototypes. BiSMAP \cite{lu2022bidirectional} utilized the source and target prototypes together to degrade those hard source samples. BAPA-Net \cite{liu2021bapa} performed prototype alignment between the mixed image and the source image.

\subsection{Domain generalization}
Domain generalization (DG) targets generalize well on other unseen domains only using the labeled source domain, where there is a domain gap between the source domain and unseen domains. Compared with the UDA task, the DG method does not need to specify the target domain during the training phase. Data generation and domain-invariant representation learning are two dominant DG methods. Data generation aims to extend the data as much as probable to cover unseen domains. For data generation, DPRC \cite{yue2019domain} generated the synthetic images with the styles of auxiliary data leveraging CycleGAN \cite{zhu2017unpaired}. FSDR \cite{huang2021fsdr} randomized images by using different domain-invariant frequencies. GLTR \cite{peng2021global} proposed and harmonized the global and local texture randomization. To extract domain-invariant features, as shown in Fig \ref{Fig:GFR} (a), some methods performed normalization or whitening on the whole features to reduce domain-specific information, such as IBN-Net \cite{pan2018two}, SW \cite{pan2019switchable}, ISW \cite{choi2021robustnet}, and SAN \cite{peng2022semantic}. Recently, PinMem \cite{kim2022pin} proposed a meta-learning framework, which memorized the domain-agnostic and class-wise distinct information to reduce the ambiguity of representation. 

\subsection{Class activation map}
Class activation map (CAM) is a technology to identify the discriminative region \cite{zhou2016learning, selvaraju2017grad, chattopadhay2018grad} by a single forward pass.  Since the visual interpretability of CAM can build the trustworthy intelligent system, CAM methods got extensively explored, such as Grad-CAM \cite{selvaraju2017grad}, Grad-CAM++ \cite{chattopadhay2018grad}. Instead improving the CAM, CAM technology is also widely used in the weakly-supervised semantic segmentation \cite{wang2020self, xu2021leveraging}, which fully leveraged discriminative localization ability of CAM. More recently, CDG \cite{du2022cross} calculated the CAM on the model trained on data from other domains as the weight to decide the feature dropout in the DG training.

\section{class-wise non-salient generalized framework}
\label{Sec:Met}
In this section, the problem statement of video generalizable semantic segmentation is first introduced. Then, the framework overview and each module of the proposed CNSG framework are individually described in detail.
\subsection{Problem statement}
Given a seen source domain \{$X_s,Y_s$\} $\in D_s$ and $K$ unseen target domains (\{$X_{t1}, Y_{t1}$\} $\in D_{t1}$, \{$X_{t2}, Y_{t2}$\} $\in D_{t2}$,...,\{$X_{tK}, Y_{tK}$\} $\in D_{tK}$), a DG model is trained using the source domain $D_s$ and then evaluates in these K unseen target domains ($D_{t1}$,...,$D_{tK}$), which aims to generalize well on all unseen target domains. $X_*$ and $Y_*$ are images and labels from different domains, respectively. The main difference between image-based and video-based DG methods is the input data. The input of image-based DG model is one image $x_*$. Compared with the image-based DG methods, the input of video-based DG methods modifies one image $x_*$ into two ($x_{*}^{t}, x_{*}^{t+1}$) or more temporally continuous frames ($x_{*}^{t-n}$,..., $x_{*}^{t-1}$, $x_{*}^{t}$, $x_{*}^{t+1}$), where the previous frames ($x_{*}^{<t+1}$) are used as the prior information to refine the segmentation of current frame $x_{*}^{t+1}$. In our case, the number of frames is 2.

\begin{figure*}[t]
\centering
\includegraphics[width=0.98\textwidth]{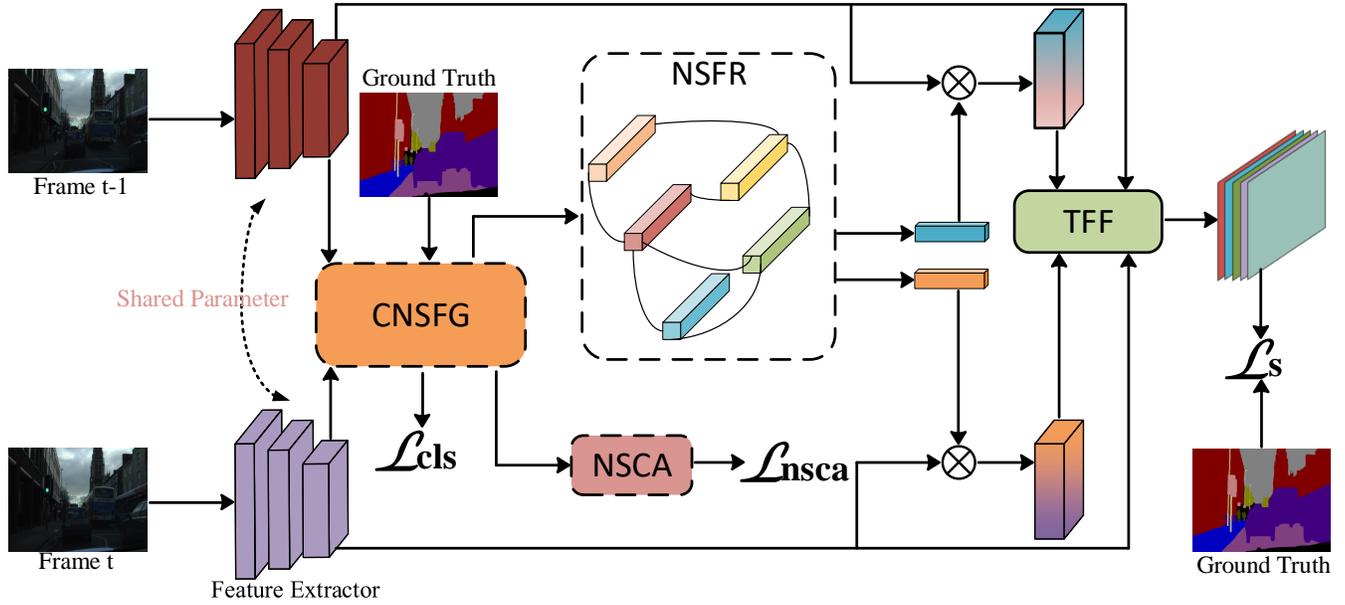}
\caption{The overview of the class-wise non-salient region generalized (CNSG) framework. CNSFG, NSFR, NSCA, TFF represent class-wise non-salient feature generalization, class-wise non-salient feature reasoning, non-salient centroid alignment, and temporal feature fusion, respectively.}
\label{Fig:Framework}
\end{figure*}
\subsection{Framework overview}
As illustrated as Fig \ref{Fig:Framework}, the proposed class-wise non-salient region generalized framework contains class-wise non-salient feature generation, class-wise non-salient feature reasoning, non-salient centroid alignment, and temporal feature fusion. The final objective of this framework can be defined as:
\begin{equation}
\mathcal{L} = \mathcal{L}_{s} + \mathcal{L}_{cls} + \mathcal{L}_{sca}
\end{equation}
where $\mathcal{L}_{s}$ is the segmentation loss. $\mathcal{L}_{cls}$ and $\mathcal{L}_{sca}$ represent the classification loss and non-salient centroid alignment loss, respectively.

\subsection{class-wise non-salient feature generation}
Since domain-invariant features helps to improve the generalizability of the model, the class-wise non-salient feature is defined, which is formed by the non-salient prototype and class activation map. The steps of class-wise non-salient feature generation contains class activation map generation, non-salient prototype extraction.
\begin{figure*}[t]
\centering
\includegraphics[width=0.98\textwidth]{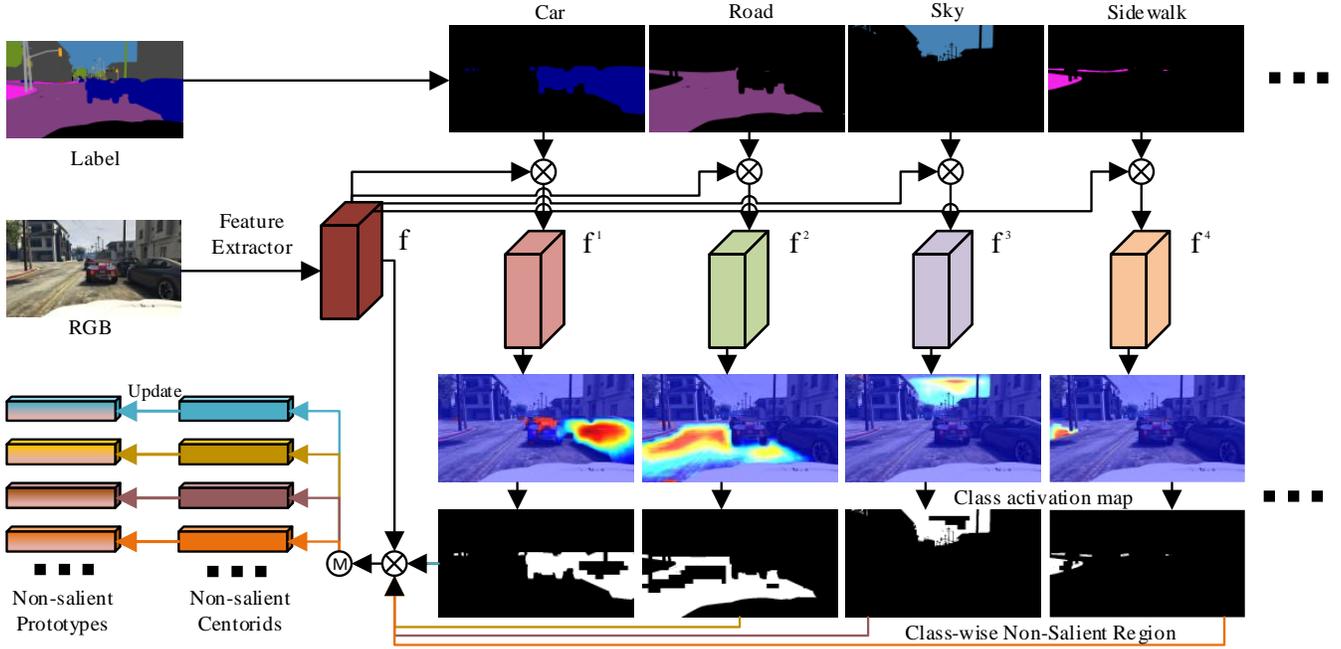}
\caption{The structure of class-wise non-salient prototype extraction. \textcircled{\scriptsize M} refers to average operation.}
\label{Fig:GFG}
\end{figure*}

\subsubsection{Class activation map generation}
To get the class activation map, a classification network as an auxiliary task is introduced. Given a training image $x_s$, the deep feature $f_o$ is extracted by the feature extractor $F(\cdot)$. Then, the $n$ class feature $f^n$ is filtered by the ground truth $y_s$, which is denoted as:

\begin{equation}
f^n =  F(x_s)\mathbb{1}(y_s^{(h,w,n)} == 1)
\label{Eq:fc}
\end{equation}
where $\mathbb{1}$ is the indicator function. The $n \in N$ class feature centroid $c^n$ is obtained by the average of the $n$ class feature $f^n$, which can be defined as:
\begin{equation}
c^n = \frac{\sum_{x_s\in X_s}\sum_{h}\sum_{w} f^n}{\sum_{x_s \in X_s}\sum_h \sum_w \mathbb{1}(y_s^{(h,w,n)}==1)}
\label{Eq:np}
\end{equation}
After that, $c^n$ is pulled into the auxiliary classifier $C^n_{cls}$ to get the prediction. The classifier loss $\mathcal{L}_{cls}$ is calculated using the cross-entropy:
\begin{equation}
\mathcal{L}_{cls} = -\sum_{i=0}^{N}y^n log(C^n_{cls}(c^n))
\end{equation}
where $\mathcal{L}_{cls}$ is designed to capture meaningful weights of the classifier. Finally, the $n$ class activation map $M^n$ is denoted as:
\begin{equation}
M^n = \sum_{k}w^n_k f_k^n
\end{equation}
where $k\in K$ is the channel of feature and $w^n_k$ represent the weight of $n$ class in $k$ channel.
\subsubsection{Non-salient prototype extraction}
As pointed out by Maxdrop \cite{park2016analysis} and RSC \cite{huang2020self} that the most predictive parts carry less domain-invariant information, the non-salient region in feature map are selected by the class activation map in our method. The non-salient class activation map is expected as more generalized regions. The $n$ class non-salient mask $M^n_{s}$ is obtained by a threshold filter, which is denoted as:  
\begin{equation}
M^n_{s}=
\begin{cases}
1& 0<M^{n^{(h,w)}} <= M^n(\alpha) \\
0& M^{n^{(h,w)}}> M^n(\alpha)
\end{cases}
\end{equation}
where $\alpha$ is a hyperparameter to represent the pixel percentage that need to be filtered. $M^n(\alpha)$ is the $J^{th}$-largest value in $M^n$, where $J = H \times W \times \alpha$. The non-salient centroid $p^n$ is calculated as the average value of $f^n$ under non-salient mask $M^n_s$, which can be denoted as:
\begin{equation}
p^n = \frac{\sum_{x_s\in X_s}\sum_{h}\sum_{w} f^n \mathbb{1}(M_s^{n^{(h,w,n)}}) == 1}{\sum_{x_s \in X_s}\sum_h \sum_w \mathbb{1}(M_s^{n^{(h,w,n)}}==1)}
\label{Eq:sp}
\end{equation}
where the $p^n$ is also the initial value of the non-salient prototype. An exponential moving average operation between the two iterations is used to update the non-salient prototype, which can be represented as:
\begin{equation}
p^n \leftarrow \lambda p^n + (1-\lambda) p^{'n}
\label{eq:ema}
\end{equation}
where $p^{'n}$ represents the non-salient centroid using the current frame by equation \ref{Eq:sp} and $\lambda$ is set to 0.9.
The non-salient prototype $p^n$ is more generalizable compared to the naive prototype obtained using the equation \ref{Eq:np}. For clarity, Fig \ref{Fig:GFG} exhibits the extraction of the non-salient prototype. Moreover, the class activation map reflects the spatial importance of each pixel, which can be seen as the prior information to enhance the model's generalizability. Hence, the class-wise non-salient feature $f_{sg}^n$ is defined as the concatenation of the non-salient prototype $p^n$ and the class activation map $M^n$, which is denoted as:
\begin{equation}
f_{sg}^n = \Psi(p^n, Flatten(M^n))
\end{equation}
where the $\Psi$ and $Flatten$ operations refer to feature concatenation and feature flattening, respectively.

\subsection{class-wise non-salient feature reasoning}
As illustrated in Section \ref{Intro}, the feature channel activation for the same classes in different domains have a large gap, which means that the model can perceive the domain information from the training images rather than only the semantic information. Different from the learning forms of Fig \ref{Fig:GFR} (a), (b), and (c), a new form is adopted to alleviate this problem, which adaptatively selects the most generalized channels and enhance features by class-wise non-salient features reasoning. Since graph convolution excels at capture node relationship and adaptively propagate information \cite{chen2019graph, li2021towards}, it is a suitable reasoning method and is used in our class-wise relationship reasoning. Given a graph $\mathcal{G}$ containing nodes $\mathcal{N}$ and edges $\mathcal{E}$, the graph convolution can be defined as:
\begin{equation}
{O_r} = \sigma(W_rf_{sg}A_r)
\end{equation}
where $O_r$, $A_r$, and $W_r$ are the output, the adjacency matrix (i.e., the relationship between nodes), and the learnable weight matrix, respectively. $\sigma$ denotes the non-linear activation function. To capture the relationship between different classes, each class non-salient feature $f^n_{sg}$ is employed as a node in the graph. A $1\times 1$ convolution layer is used to get the adjacent matrix $A_r$ like GloRe \cite{chen2019graph}. Meanwhile, the relationship reasoning also conducts Laplacian matrix smoothing $(I - A_r)$ by a residual sum between the identity matrix $I$ and the adjacent matrix $A_r$ to better propagate the node features over the graph. Therefore, the graph relationship reasoning can be rewritten as:
\begin{equation}
{O_r} = \sigma(W_rf_{sg}(I-A_r))
\end{equation}
After that, another $1\times 1$ convolution $W_a$ is utilized to match the channel dimension of original feature $f_o$. The reconstructed feature $\hat{f}$ can be defined as:
\begin{equation}
\bar{f} = \delta(W_aO_r)*f_o+f_o
\end{equation}
where $\delta$ is the sigmoid operation. 

\subsection{Temporal feature fusion}
Considering the temporal frames, the $t$ frame final feature $f_t$ can be concatenated from the high-level feature $f_{t_{h}}$, the low-level feature $f_{t_l}$ and the reconstructed feature $\bar{f}_t$, which can be defined as:
\begin{equation}
\label{finalf} 
f_t= \Psi(U(f_{t_{h}}),f_{t_{l}}, U(\bar{f}_t))
\end{equation}
where the original feature $f_{t_{o}}$ is pulled into the $ASPP$ module \cite{chen2017deeplab} to aggregate multi-scale context, where $f_{t_{h}}=ASPP(f_{t_o})$. The low-level feature $f_{t_l}$ is the feature of stage $1$ in backbone network. $U$ refers to the unsample operation to match the dimension of $f_{t_l}$. The temporal fused prediction $P_{fuse}$ is obtained by concatenating predictions of two frames, which can be denoted as:
\begin{equation}
\label{tff} 
P_{fuse} = C_{fuse}(\Psi(C(f_t), \mathcal{W}(C(f_{t-1}),\mathcal{F})))
\end{equation}
where $\mathcal{W}$ refers to the warpping operation and $\mathcal{F}$ is the optical flow estimated by FlowNet-V2 \cite{ilg2017flownet}. $C_{fuse}$ is the classifier for the temporal fused prediction. Finally, a cross-entropy function is leveraged as the segmentation loss:
\begin{equation}
\label{segloss} 
\mathcal{L}_{s}=-\sum_{h,w}\sum_{n\in N}y_s^{(h,w,n)}log(P_{fuse}^{(h,w)}))
\end{equation}

\subsection{non-salient centroid alignment}
To alleviate the prediction inconsistency between two frames, a naive method is to align class-wise features in different frames, such as the class centroid. As illustrated in the above paragraph, since the non-salient centroid is generated by the more generalized region, performing the non-salient centroid alignment may be a better choice than the naive centroid. Therefore, an inter-frame non-salient centroid alignment loss is proposed, which can be described as:
\begin{equation}
\mathcal{L}_{sca} = \frac{1}{N}\sum_{n=0}^N |p^n_{t-1}-p^n_{t}| 
\end{equation}
where the non-salient centroid is calculated by equation \ref{Eq:sp}. The effect of the alignment loss is two fold. First, the inter-frame non-salient centroid alignment encouages the alignment of generalized features between different frames rather than the global features, which enhance the generalizability of the model. Second, the inter-frame feature alignment alleviate the prediction inconsistency. 

\section{Experiment}
In this section, extensive experiments are conducted to verify the superiority of our CNSG framework in the VGSS and IGSS tasks, which include qualitative and quantitative comparison, ablation studies and so on. Additionally, the adopted datasets and experimental detail are introduced.

\label{Sec:Exp}
\subsection{Dataset}
Seven datasets are used in both IGSS and VGSS tasks, which include four real-world datasets (Cityscapes \cite{cordts2016cityscapes}, CamVid \cite{brostow2008segmentation}, Mapillary \cite{neuhold2017mapillary}, BDD100K \cite{yu2020bdd100k}) and three synthetic datasets (VIPER \cite{richter2017playing}, GTAV \cite{richter2016playing}, SYNTHIA \cite{ros2016synthia}).
 
\subsubsection{datasets for the VGSS task}
Four datasets are utilized in the VGSS task. For synthetic datasets, the VIPER dataset is a synthetic dataset on the urban scene containing over 25000 video frames with the FHD resolution (1920 $\times$ 1080) in different environmental conditions, which is captured by the computer game Grand Theft Auto V. The SYNTHIA-Seq dataset, the sub-dataset of the SYNTHIA, has 8500 video frames with eight views, where the images of six views and the images of the other two views are split into the training and validation sets, respectively. For real-world datasets, 5950 training and 1000 validation video frames with the resolution of $2048 \times 1024$ are adopted in the Cityscapes-Seq dataset. The CamVid dataset has 366 and 101 video frames with the resolution of $960 \times 720$ for the training and validation sets respectively. 
\subsubsection{datasets for the IGSS task}
Five datasets are adopted in the IGSS task. For synthetic datasets, the GTAV dataset collected over 25000 images with a high resolution of $1914 \times 1052$. The SYNTHIA-Rand set is the subset of the SYNTHIA dataset and consists of 9400 $1280\times 760$ images with different views. For real-world datasets, the BDD100K dataset is a large driving dataset, where 7000 $1280 \times 720$ training and 1000 validation images are used in the IGSS task. The Mapillary dataset covers six continents and consists of 18000 training and 2000 validation images, where the resolution of the images is at least FHD resolution. The Cityscapes dataset is widely used in the street understanding task, which contains 2975 training and 500 validation images with the resolution of $2048 \times 1024$. 

\subsection{Implementation detail}
ResNet-50 \cite{he2016deep}, ShuffleNet-V2 \cite{ma2018shufflenet}, and MobileNet-V2 \cite{sandler2018mobilenetv2} are adopted as the backbone networks and the pre-trained model on the ImageNet is used as the initial model. The segmentation model adopt a Stochasitc Gradient Descent (SGD) optimizer using Pytorch library \cite{choi2021robustnet}, where the initial learning rate, weight decay, and momentum are set to 0.01, 5e-4, and 0.9, respectively. Mean Intersection over Union (MIoU) of categories are reported as the performance metric for comparison. Like ISW \cite{choi2021robustnet}, several photometric transformations, such as Gaussian blurring and color jittering, are used to avoid source overfitting. The hyperparameter of the filter for non-salient centroid generation is set to 0.3. For the VGSS task, all video frames are resized to 768$\times$384. For the IGSS task, all input images are resized to 768$\times$768 like other methods (ISW \cite{choi2021robustnet} and PinMem \cite{kim2022pin}).

\begin{table*}[t]
\caption{Quantitative comparisons for the VGSS task on ResNet-50, MobileNet-V2, and ShuffleNet-V2 backbones. The best and second-best performances are represented by \textbf{bold} and \uline{underline}, respectively. $\rightarrow$ refers to ``generalize to''. Avg refers to the average mIoU on different evaluation datasets.}
\vspace{-2mm}
\label{tab:VDG performance}
\centering{}\resizebox{1.0\textwidth}{!}{%
\begin{tabular}{ccc|ccc|ccc|ccc|cccccccc}
\toprule 
\multirow{2}{*}{Methods}& \multirow{2}{*}{Model} & \multirow{2}{*}{Avg}& \multicolumn{3}{c}{ VIPER ($V_s$)$\rightarrow$} &  \multicolumn{3}{c}{ Synthia-Seq ($S_s$)$\rightarrow$} &   \multicolumn{3}{c}{CamVid ($CV_s$)$\rightarrow$} &  \multicolumn{3}{c}{Cityscapes-Seq ($C_s$)$\rightarrow$}  \\
 && & $\rightarrow$$S_s$ & $\rightarrow$$C_s$ & $\rightarrow$$CV_s$ & $\rightarrow$$V_s$ & $\rightarrow$$C_s$ & $\rightarrow$$CV_s$ & $\rightarrow$$V_s$ & $\rightarrow$$C_s$ &  $\rightarrow$$S_s$ & $\rightarrow$$V_s$ & $\rightarrow$$CV_s$ & $\rightarrow$$S_s$  \tabularnewline 
\hline 
\noalign{\vskip0.1cm}
IBN \cite{pan2018two} & ResNet-50 &36.5 & 36.3 & 29.6 & 33.5 & 21.1 & 40.0 & \uline{45.7} & 21.6 & 44.6 & 39.6 & {23.0} & 60.2 & 40.9 \tabularnewline
SW \cite{pan2019switchable} & ResNet-50& \uline{38.4}& \textbf{45.2} & 28.9 & 35.2 & 22.6  & \uline{41.0} & \uline{45.7} & {22.9} & {47.7} &  {40.2} & 21.9 & {60.8} & \uline{43.8}\tabularnewline
ISW \cite{choi2021robustnet} & ResNet-50 & 37.7 & \uline{42.3} & \uline{34.0} &{38.6}& \uline{24.4} & 40.1 &  44.6 & 21.1& 43.7 & 39.2 & 20.8 & 60.1 & 40.5  \tabularnewline
PinMem \cite{kim2022pin} & ResNet-50&\uline{38.4}&40.6  &30.8 &\textbf{43.2} & 21.4 &37.5 &32.1 & \uline{23.9} &\textbf{56.0} &\uline{43.6} & \uline{29.9} & \uline{61.1} & 40.2 
\tabularnewline
CNSG (Ours)  & ResNet-50 & \textbf{41.8} & 41.1 & \textbf{36.0} &\uline{39.6}&\textbf{26.1}&\textbf{42.1}&\textbf{47.8}&\textbf{27.2}&\uline{52.7}&\textbf{43.7}&\textbf{33.1}&\textbf{61.7}&\textbf{50.9}  & \tabularnewline[0.1cm]
\toprule 
IBN \cite{pan2018two} & MobileNet-V2 &33.6 & 35.8 & {26.2} & 34.2 & \uline{21.6} & \uline{35.9} & 40.0 & 18.5& 43.9  & 37.4& 17.4 & 55.5 & 36.8 \tabularnewline
SW \cite{pan2019switchable} & MobileNet-V2& {34.4} & \uline{38.3} & 23.5 & \uline{37.5} & 20.0  & 34.7 & \uline{42.6} & {21.4}& {46.2}& \textbf{40.4} & 15.3 & \uline{57.7} & 34.6\tabularnewline
ISW \cite{choi2021robustnet} & MobileNet-V2 & 34.1& 35.5 & 25.6 & 34.6 & \uline{21.6} & \uline{35.9} & 40.8 & 18.7 & 43.7 & 36.3 & 19.2 & \textbf{58.6} & {39.1} \tabularnewline
PinMem \cite{kim2022pin} & MobileNet-V2 & \uline{35.3} & 37.6 &\uline{27.1} &\textbf{43.3}& 18.7 & 24.9 & 31.0 & \uline{23.8} & \textbf{51.5} & 36.2 &\uline{29.4} & 56.0 & \uline{43.9} \tabularnewline
CNSG (Ours)  & MobileNet-V2& \textbf{38.5} & \textbf{38.8} & \textbf{31.9} &34.3&\textbf{26.2}&\textbf{39.8}&\textbf{45.7}&\textbf{25.1}&\uline{50.1}&\uline{39.1}&\textbf{30.4}&55.3&\textbf{45.6}  & \tabularnewline[0.1cm]
\toprule 
IBN \cite{pan2018two} & ShuffleNet-V2 &31.7 &34.5 &{23.3}& 27.9& 23.3 &32.5 &43.1& 19.5& 38.1& 31.1 &16.9 &54.6 &35.5 
 \tabularnewline
SW \cite{pan2019switchable} & ShuffleNet-V2& 31.6 &\uline{35.4}&21.8 &\uline{33.6}& 21.7 &\uline{33.5} &40.8&{19.9}& \uline{39.2}& {32.9} &15.2& 53.5 &32.1 
\tabularnewline
ISW \cite{choi2021robustnet} & ShuffleNet-V2 & {32.5} &35.1 &\uline{24.4} &28.9 &22.1 &32.1& \uline{43.6} &18.9 &38.1 &30.9 & {22.2}& {55.5}& {37.9}
 \tabularnewline
PinMem \cite{kim2022pin} & ShuffleNet-V2 & \uline{33.0} &32.3 &\textbf{28.5} &29.8 & 17.2& 23.7& {26.7} & \uline{22.7} & \textbf{52.1} & \textbf{37.2} & \uline{27.4}& \textbf{56.5}& \uline{42.0}
 \tabularnewline
CNSG (Ours)  & ShuffleNet-V2 & \textbf{36.6} & \textbf{36.6} & \textbf{28.5} &\textbf{36.3}&\textbf{23.8}&\textbf{35.2}&\textbf{44.8}&\textbf{24.2}&\uline{46.6}&\uline{36.8}&\textbf{27.9}&\uline{56.3}&\textbf{42.5}  & \tabularnewline[0.1cm]
\toprule
\noalign{\vskip0.1cm}
\end{tabular}}
\vspace{-1mm}
\end{table*}

\begin{table*}[t]
\caption{Quantitative comparisons for the IGSS task on ResNet-50 backbone. The best and second-best performances are represented by \textbf{bold} and \uline{underline}, respectively. $\rightarrow$ refers to ``generalize to''. Avg refers to the average mIoU on different evaluation datasets.}
\vspace{-2mm}
\label{tab:DG performance}
\centering{}\resizebox{1.0\textwidth}{!}{%
\begin{tabular}{ccc|cccc|cccc|cccc|cccc|cccccccccc}
\toprule 
\multirow{2}{*}{Methods}& \multirow{2}{*}{Model} & \multirow{2}{*}{Avg}& \multicolumn{4}{c}{GTAV (G)$\rightarrow$} &  \multicolumn{4}{c}{SYNTHIA (S)$\rightarrow$} &   \multicolumn{4}{c}{ Cityscapes (C)$\rightarrow$} &  \multicolumn{4}{c}{BDD (B)$\rightarrow$} &  \multicolumn{4}{c}{Mapillary (M)$\rightarrow$} \\
 && & $\rightarrow$C & $\rightarrow$B & $\rightarrow$M & $\rightarrow$S & $\rightarrow$C & $\rightarrow$B & $\rightarrow$M & $\rightarrow$G &  $\rightarrow$B & $\rightarrow$M & $\rightarrow$G & $\rightarrow$S &  $\rightarrow$G & $\rightarrow$S & $\rightarrow$C & $\rightarrow$M &  $\rightarrow$G & $\rightarrow$S & $\rightarrow$C & $\rightarrow$B  \tabularnewline 
\hline 
\noalign{\vskip0.1cm}
IBN \cite{pan2018two} & ResNet-50 &34.2 & 33.9 & 32.3 & 37.8 & 27.9 &  32.0 & 30.6 & 32.2 & 26.9  & 48.6 & 57.0 & 45.1 & 26.1 & 29.0 & 25.4& 41.1 & 26.6 &  30.7 & 27.0 & 42.8 & 31.0\tabularnewline
SW \cite{pan2019switchable} & ResNet-50& 32.2 & 29.9 & 27.5 & 29.7 & 27.6  & 28.2 & 27.1 & 26.3 & 26.5 &  48.5 & 55.8 & 44.9 & 26.1&  27.7 & 25.4 & 40.9 & 25.8 & 28.5 & 27.4 & 40.7 & 30.5\tabularnewline
DRPC \cite{yue2019domain} & ResNet-50 & 35.8 & 37.4 & 32.1 & 34.1 & 28.1 &35.7 & 31.5 & 32.7 & 28.8 &  49.9 & 56.3 & 45.6 & 26.6 & 33.2 & 29.8 & 41.3 & 31.9 &  33.0 & 29.6 & 46.2 & 32.9\tabularnewline
GTR \cite{peng2021global}& ResNet-50 & 36.1 & 37.5 & 33.8 & 34.5 & 28.2 & 36.8 & {32.0} & {32.9} & 28.0 & 50.8 & 57.2 & \uline{45.8} & 26.5 & {33.3} & 30.6 & 42.6 & 30.7  & 32.9 & 30.3 & 45.8 & 32.6 \tabularnewline
ISW \cite{choi2021robustnet} & ResNet-50 & 36.4 & 36.6 & 35.2 & 40.3 & 28.3 &  35.8 & 31.6 & 30.8 & 27.7  & 50.7 & \uline{58.6} & 45.0 & 26.2 &  32.7 & 30.5 & 43.5 & 31.6 & 33.4 & 30.2 & 46.4 & 32.6\tabularnewline
SAN \cite{peng2022semantic} & ResNet-50 & {38.5} &  {39.8} & \uline{37.3} & \uline{41.9} & \uline{30.8}  & \uline{38.9} & \textbf{35.2} & \textbf{34.5} & {29.2} &   \textbf{53.0} & \textbf{59.8} & \textbf{47.3} & {28.3}  & \uline{34.8} & \uline{31.8} & {44.9} &{33.2}&  {34.0} & \uline{31.6} & {48.7} & {34.6} \tabularnewline[0.1cm]
PinMem \cite{kim2022pin} & ResNet-50& \uline{41.0} & \uline{41.2} & 35.2 & 39.4 & 28.9 & 38.2 & \uline{32.3} & \uline{33.9} & \textbf{32.1} & 50.6 & 57.9 &45.1 & \uline{29.4} & \textbf{42.4} & 29.1 & \uline{54.8} & \uline{51.0} &\uline{44.1} & 30.8 & \uline{55.9} &\uline{47.6} \tabularnewline[0.1cm]
CNSG (Ours)  & ResNet-50 & \textbf{42.2} & \textbf{42.6} & \textbf{37.9} &\textbf{42.0}&\textbf{33.1}&\textbf{39.5}&30.0&32.3&\uline{29.7}&\uline{50.9}&57.8&45.3&\textbf{30.5}&30.6&\textbf{33.6}&\textbf{57.4}& \textbf{56.2} & \textbf{49.7}  & \textbf{34.6}& \textbf{60.0}& \textbf{51.2}  & \tabularnewline[0.1cm]
\hline 
\noalign{\vskip0.1cm}
\end{tabular}}
\vspace{-1mm}
\end{table*}

\begin{table}[t]
    \centering
    \tabcolsep=0.09cm
    \caption{Ablation studies on proposed component containing NSFR and NSCA. The model is trained on the Cityscapes-Seq with ResNet-50 backbone network.}
    \small
    \begin{tabular}{l|cc|cccc}
         \bottomrule
         Method &  NSFR & NSCA   & $\rightarrow$ $V_s$ &$\rightarrow$$CV_s$&$\rightarrow$$S_s$ & Avg \\
	   \cline{1-7}
        Baseline &&  & 28.2 &\uline{60.8} &40.0 &43.0  \\
		NSCA  &  & \checkmark   &32.6& 57.4 &\uline{50.0} 
& 46.7 \\
        NSFR &\checkmark&   &\textbf{34.1}& 60.0 & {49.5} &\uline{47.9}  \\	
        NSFR $+$ NSCA &\checkmark&   \checkmark & \uline{33.1} &\textbf{61.7}& \textbf{50.9}  & \textbf{48.6}\\
         \toprule
    \end{tabular}
\label{table:ABC}
\end{table}

\begin{table}[t]
    \centering
    \tabcolsep=0.09cm
    \caption{Ablation studies on internet component of class-wise non-salient fetaure. NSP, NP, and CAM are the non-salient prototype, naive prototype and the class activation map, respectively.}

    \small
    \begin{tabular}{l|ccc|cccc}
         \bottomrule
         Method &  NP & CAM & NSP  & $\rightarrow$ $V_s$ &$\rightarrow$$CV_s$&$\rightarrow$$S_s$ & Avg \\
	   \cline{1-8}
        Baseline &&  && 28.2 &\uline{60.8} &40.0 &43.0  \\
        NP &\checkmark&  & &{33.1} & {58.4} &44.8 & {45.4}  \\
			CAM  &  & \checkmark&   &32.7 &60.4 &43.7 & 45.6 
\\
        NSP &&  & \checkmark&\uline{33.8} & {57.8} &\uline{48.2} & \uline{46.6}  \\
		NP + CAM & \checkmark & \checkmark& & 32.8& \textbf{61.2}& 45.0 &46.3 
\\
		NSP + CAM & & \checkmark &\checkmark  & \textbf{34.1}& 60.0 & \textbf{49.5} & \textbf{47.9}
\\
         \toprule
    \end{tabular}
\label{table:ABC_In}
\end{table}

\begin{table}[t]
    \centering
    \tabcolsep=0.09cm
    \caption{Ablation studies on internet component of inter-frame non-salient centroid alignment. NSCA and NCA represent the non-salient centroid and naive centroid alignments, respectively.}
    \small
    \begin{tabular}{l|cc|cccc}
         \bottomrule
         Method &  NSCA & NCA   & $\rightarrow$ $V_s$ &$\rightarrow$$CV_s$&$\rightarrow$$S_s$ & Avg \\
	   \cline{1-7}
        Baseline &&  & 28.2 &\textbf{60.8} &40.0 &43.0  \\
		
        NCA &\checkmark&   &\textbf{32.8}& {56.9} & \uline{45.4} &\uline{45.0}  \\	
			NSCA  &  & \checkmark   &\uline{32.6}& \uline{57.4} &\textbf{50.0} & \textbf{46.7} \\
         \toprule
    \end{tabular}
\label{table:ABC_InCA}
\end{table}

\begin{figure*}[t]
\centering
\includegraphics[width=0.98\textwidth]{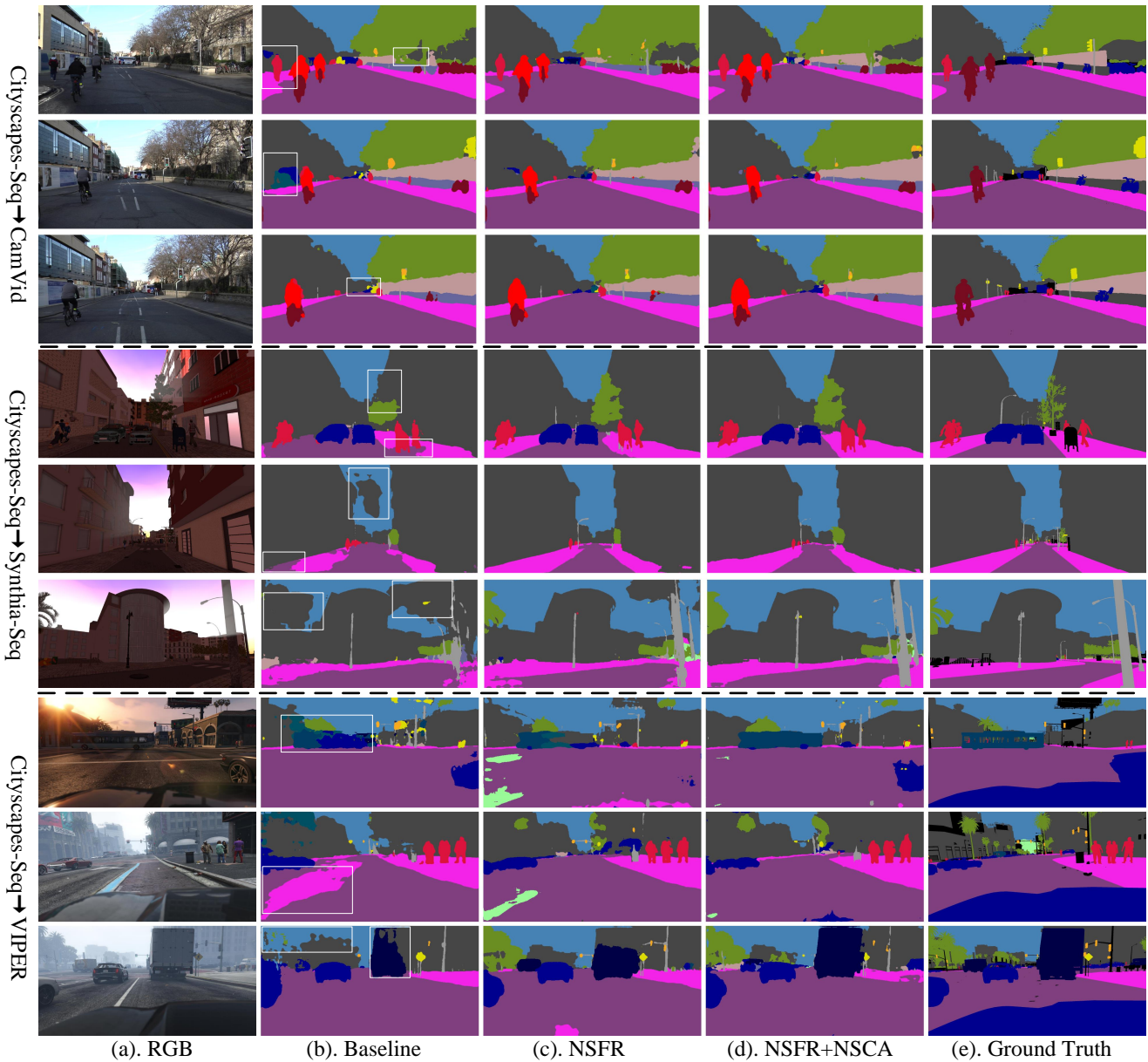}
\caption{Result visualization comparison of different models on distinct domains. From left to right: input image, baseline results, the results with NSFR, the results with both proposed components (NSFR + NSCA), and ground truth. Best viewed in color. White boxs show the segmentation error in the baseline model.}
\label{Fig:Visualization}
\end{figure*}

\subsection{Comparisons with state-of-the-art methods}
We validate the effectiveness of our proposed CNSG framework in the VGSS and IGSS tasks.
\subsubsection{Comparisons on the VGSS task}
In the VGSS task, the video dataset VIPER ($V_s$), Synthia-Seq ($S_s$), CamVid ($CV_s$), Cityscapes-Seq ($C_s$) is taken one by one in turn as the source domain for training. In other word, four evaluation settings are conducted, which contains $V_s\rightarrow \{S_s, C_s, CV_s\}$, $S_s \rightarrow \{V_s, C_s, CV_s\}$, $C_s \rightarrow \{V_s, CV_s, S_s\}$, and $CV_s \rightarrow C_s, V_s, S_s$. The left of the $\rightarrow$ is the source domain for training, and the right of the $\rightarrow$ is the target domain for evaluation. One model is selected to evaluate all target domains instead of using multiple models like DPRC \cite{yue2019domain}. Performance comparison are reported in Table \ref{tab:VDG performance}. In the ResNet-50 backbone, our method achieves 41.8\% in terms of average mIoU and gets 9 best results in 12 evaluation settings. A 3.4\% improvement in average mIoU shows the superiority of our approach compared to the second-best method. In the MobileNet-V2 backbone, our approach outperforms all the state-of-the-art methods with a significant improvement of at least 3.2\% in average mIoU. 9 performances in 12 evaluation settings in the ShuffleNet-V2 backbone achieve the best. Compared to the second-best method, the result of our framework is improved to 36.6\% and has a clear increase of 3.6\% in terms of average MIoU. These experiments and discussions indicate that the segmentation quality is enhanced by the proposed class-wise relationship reasoning and non-salient centroid alignment.
\par The segmentation visualizations of different models (the baseline model, the model with NSFR, and the model with NSFR + NSCA) on distinct domains ($C_s\rightarrow CV_s$, $C_s\rightarrow S_s$, $C_s\rightarrow V_s$) are provided for comparison in Fig \ref{Fig:Visualization}. The NSFR model and the final model exhibit better results than the baseline model and the final model show less segmentation error than the NSFR model, demonstrating that our framework alleviates class confusion problem. For example, in the first row of $C_s \rightarrow V_s$, the bus shape in our final model is closer to the ground truth.
\subsubsection{Comparisons on the IGSS task}
In addition, we report the performance in the IGSS task in Table \ref{tab:DG performance}, where the model is trained on Mapillary (M), GTAV (G), Cityscapes (C), BDD100K (B), and SYNTHIA (S) in turn similar to the VGSS task. Correspondingly, experiments with five evaluation settings are performed, i.e., G$\rightarrow$ \{C, B, M, S\}, S$\rightarrow$ \{G, C, B, M\}, C $\rightarrow$ \{G, S, B, M\}, B $\rightarrow$ \{G, C, S, M\}, M $\rightarrow$\{G, S, C, B\}. 13 best and 2 second-best performances in our framework show the CNSG framework achieves state-of-the-art performance. Our approach improves the performance by 1.2\% in terms of average mIoU compared to the second-best method (i.e., PinMem \cite{kim2022pin}). These experiments verify that our proposed method also works for the IGSS task and outperforms other state-of-the-art methods.
\subsection{Ablation study}
To demonstrate the effectiveness of the proposed components, ablation studies are conducted. NSFR and NSCA are ablation terms for evaluation. As shown in Table \ref{table:ABC}, the performance of the model trained on Cityscapes-Seq on the ResNet-50 backbone is reported and the rest three datasets are evaluated. The baseline model denotes the model only with temporal feature fusion. Our NSCA module achieves 46.7\% in terms of average mIoU with 3.7\% improvement compared to the baseline. Furthermore, NSFR gets 47.9\% in terms of average mIoU. As last, the performance of the final model is improved to 48.6\% average mIoU and the performance improvement is obvious compared to the baseline. The proposed modules contribute to enhancing the generalizability of the model.

\par Moreover, to validate the effectiveness of the non-salient region, ablation studies about sub-components of the proposed modules are conducted. First, as mentioned in Section \ref{Sec:Met}. C, the class-wise non-salient feature is concatenated by the non-salient prototype and the class activation map. Table \ref{table:ABC_In} shows the effect of the above sub-components. The model with the non-salient prototype has a gain of 1.2\% in terms of average performance compared to the mode with the naive prototype. Meanwhile, the NSP $+$ CAM model achieves 47.9\% in average mIoU and outperforms the NP $+$ CAM model by 1.3\% average mIoU, which shows that the features in the non-salient region carry more generalized information. Second, as shown in Table \ref{table:ABC_InCA}, the proposed non-salient centroid alignment achieves an improvement of 1.7\% in average mIoU over the naive centroid alignment. These experiments demonstrate the effectiveness of the proposed components and non-salient region.
\begin{figure}[t]
\centering
\includegraphics[width=0.48\textwidth]{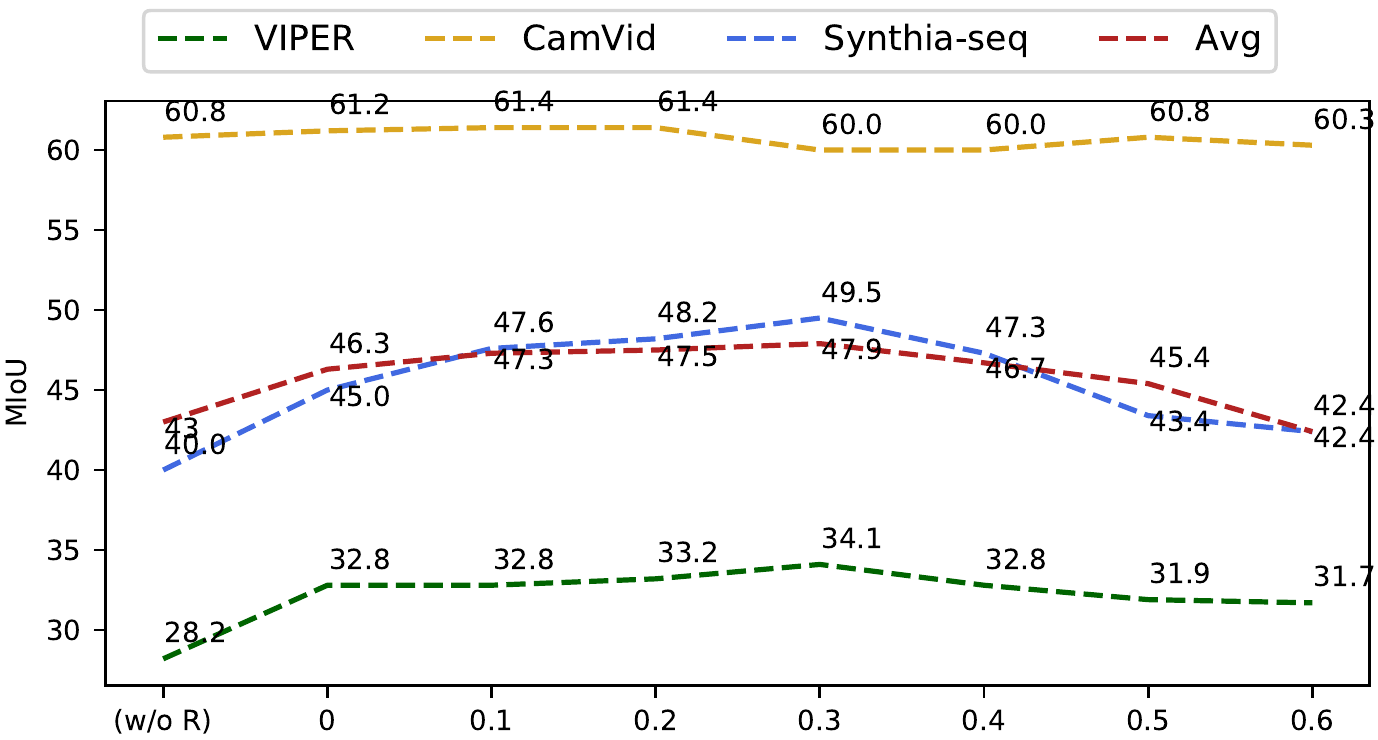}
\caption{The hyperparameter evaluation of deciding the non-salient region. (w/o R) denotes the results without NSFR.}
\label{Fig:Parameter}
\end{figure}

\subsection{non-salient region hyperparameter evaluation}
The hyperparameter of deciding the non-salient region is also important, where the hyperparameter $\alpha$ represents the filter percentage. For example, 30\% pixels in the feature map will be filtered when the hyperparameter equals $0.3$. As exhibited in Fig \ref{Fig:Parameter}, the average performance achieves the best when the hyperparameter $\alpha$ equals 0.3. Meanwhile, it can be found that the performances reasonably  increase and then decrease. First, the performance without NSFR shows the worst. Then, the model with feature reasoning but without the non-salient region filter ($\alpha = 0$) improves the generalized performance, which verifies the effectiveness of the feature reasoning. Next, since the feature of most salient regions that provides less generalizable information is filtered, the performance increases when the $\alpha$ increases. Finally, features carrying domain-invariant information are filtered, which leads to decreased performances when the $\alpha$ increases further.

\section{Conclusion}
In this paper, we introduce a new important task due to the continuous attribute of the real world, video generalizable semantic segmentation (VGSS) that considers both continuous data and domain generalization. To the best of our knowledge, this task has not been studied in existing research. For this new task, we propose a class-wise non-salient region generalized (CNSG) framework. Specifically, we first define the class-wise non-salient feature, which describes features of the class-wise non-salient region that carry more generalizable information. Then, we propose a class-wise non-salient feature reasoning strategy to adaptatively select and enhance the most generalizable channels. Finally, we propose an inter-frame non-salient centroid alignment loss to alleviate the predicted inconsistent problem under the domain generalization setting. Furthermore, we extend our method to the image-based generalizable semantic segmentation (IGSS) task. Extensive results on both VGSS and IGSS tasks show the effectiveness of our CNSG framework.
\label{Sec:Coc}

\bibliography{ref} 
\bibliographystyle{IEEEtran} 

\end{document}